\documentclass[letterpaper]{article} 
\usepackage[]{aaai25}  
\usepackage{enumitem}
\usepackage{adjustbox}
\usepackage{times}  
\usepackage{helvet}  
\usepackage{courier}  
\usepackage[hyphens]{url}  
\usepackage{graphicx} 
\usepackage{natbib}  
\usepackage{caption} 
\usepackage{algorithm}
\usepackage{algorithmic}
\usepackage{amsmath}
\usepackage{amsfonts}
\usepackage{mathtools}
\usepackage{amssymb}
\newcommand{\vx}{\mathbf{x}}
\newcommand{\vy}{\mathbf{y}}
\newcommand{\norm}[1]{\left\lVert#1\right\rVert}
\usepackage{newfloat}
\usepackage{listings}
\usepackage{threeparttable}

\usepackage{url}            
\usepackage{booktabs}       
\usepackage{amsfonts}       
\usepackage{nicefrac}       
\usepackage{microtype}      
\usepackage{xcolor}         

\usepackage{times}
\usepackage{epsfig}
\usepackage{graphicx}
\usepackage{amsmath}
\usepackage{amssymb}

\usepackage{xcolor}
\usepackage{algorithm}
\usepackage{listings}
\usepackage[british,american]{babel}
\usepackage{enumitem}
\usepackage{multirow}

\usepackage{textpos}  
\usepackage{float}
\usepackage{siunitx}
\usepackage{caption}
\usepackage{subcaption}
\usepackage{amsthm}

\usepackage[most]{tcolorbox}
\usepackage{listings}
\usepackage{pifont}
\usepackage{xspace}
\usepackage{soul}
\usepackage{etoc}
\usepackage{colortbl}
\usepackage{algorithm}
\usepackage{xcolor}
\usepackage{url}
\usepackage{csquotes}
\usepackage{graphicx}
\usepackage{amsmath}
\usepackage{mathtools}
\usepackage{amssymb}
\usepackage{amsfonts}
\usepackage{bm}
\usepackage{booktabs}
\usepackage{multirow}
\usepackage{enumitem}
\usepackage{algorithm}
\usepackage{xcolor}
\usepackage{rotating}
\usepackage{makecell}
\usepackage{listings}
\usepackage{colortbl}

\usepackage{url}
\usepackage{graphicx}
\usepackage{csquotes}
\usepackage{graphicx}
\usepackage{amsmath}
\usepackage{mathtools}
\usepackage{amssymb}
\usepackage{amsfonts}
\usepackage{bm}
\usepackage{booktabs}
\usepackage{multirow}
\usepackage{enumitem}
\usepackage{algorithm}
\usepackage{xcolor}
\usepackage{rotating}
\usepackage{makecell}
\usepackage{listings}
\usepackage{colortbl}
\usepackage{pifont}
\usepackage{xspace}
\usepackage{soul}
\usepackage{etoc}
\newtheorem{theorem}{Theorem}

\DeclareCaptionStyle{ruled}{labelfont=normalfont,labelsep=colon,strut=off} 
\floatstyle{ruled}
\newfloat{listing}{tb}{lst}{}
\floatname{listing}{Listing}
\definecolor{citecolor}{HTML}{00693e}
\definecolor{revisedcolor}{HTML}{9F363A}

\title{Scaled Supervision is an Implicit Lipschitz Regularizer}
\author{
    Zhongyu Ouyang, Chunhui Zhang, Yaning Jia, Soroush Vosoughi\thanks{Corresponding author}
}
\affiliations{
    Dartmouth College, Hanover, NH 03755\\
    \{zhongyu.ouyang.gr, chunhui.zhang.gr, yaning.jia.gr, soroush.vosoughi\}@dartmouth.edu
}

\newcommand{\answerYes}[1]{\textcolor{blue}{#1}} 
\newcommand{\answerNo}[1]{\textcolor{teal}{#1}} 
\newcommand{\answerNA}[1]{\textcolor{gray}{#1}} 
 
\usepackage{xcolor}
\definecolor{meta-blue}{HTML}{1877F2}
\begin{document}

\maketitle
\begin{abstract}
In modern social media, recommender systems (RecSys) rely on the click-through rate (CTR) as the standard metric to evaluate user engagement. CTR prediction is traditionally framed as a binary classification task to predict whether a user will interact with a given item. 
However, this approach overlooks the complexity of real-world social modeling, where user, item, and their interactive features change dynamically in fast-paced online environments. This dynamic nature often leads to model instability, reflected in overfitting short-term fluctuations rather than higher-level interactive patterns.  
While overfitting calls for more scaled and refined supervisions, current solutions often rely on binary labels that overly simplify fine-grained user preferences through the thresholding process, which significantly reduces the richness of the supervision.
Therefore, we aim to alleviate the overfitting problem by increasing the supervision bandwidth in CTR training.
Specifically, \textit{(i)} theoretically, we formulate the impact of fine-grained preferences on model stability as a Lipschitz constrain;
\textit{(ii)} empirically, we discover that scaling the supervision bandwidth can act as an \textit{implicit} Lipschitz regularizer, stably optimizing existing CTR models to achieve better generalizability.
Extensive experiments show that this scaled supervision significantly and consistently improves the optimization process and the performance of existing CTR models, even without the need for additional hyperparameter tuning\footnote{\url{https://github.com/zyouyang/ImpLipReg_CTR.git}}. 
\end{abstract}

\section*{Introduction}
In modern social media, recommender systems are ubiquitous in online applications and have redefined the user experience in product recommendation on e-commerce platforms~\cite{wang2021dcn, schafer1999recommender}, personalized modelling~\cite{gomez2015netflix, van2013deep, wu2022graphbert}, and friend recommendation on social media platforms~\cite{ma2008sorec, jamali2010matrix, fan2019graph, wen2022opioid, ouyang2024symbol, ouyang2024cf}. 
As an important type of RecSys, click-through rate (CTR) models predict users' interaction probability, which help developers improve user engagement and recommendation accuracy in social media. 
CTR models specifically incorporate contextual features (e.g., user's demographic information, item descriptions, or transaction timestamps between a user-item interaction) when making interaction prediction.
The problem is generally formulated as a binary classification task.

While the above CTR model training pipeline is commonly adopted, it often overlooks the underlying dynamics in the embeddings learned from the recommendation environment.
Interaction prediction is delicately influenced by various factors, including recent user behavior, trends, and contextual information, all of which can shift rapidly over time.
This dynamic nature poses a significant challenge for CTR models - they may struggle to keep pace with the continual changes as the binary classification formulation oversimplifies the structure of preferences.
As a result, the models could overfit short-term fluctuations in the data, failing to capture general patterns that are representative of long-term interactions and impairing its stability to input noises.

The Lipschitz constant is commonly adopted to assess the stability of a model.
A smaller Lipschitz constant indicates that the model’s output is less prone to change drastically in response to minor variations in the input, indicating better stability.
Model stability is essential for practical CTR prediction, given the fast-changing nature of inputs in RecSys.
Since there are massive studies~\cite{shi2022efficiently, wang2024on} that acknowledge accurately estimating the Lipschitz constant for a deep model is challenging due to the complexity and diversity of the feature construction layers involved, directly estimating a CTR model's Lipschitz constant is impractical.
Alternatively, we study the \textit{theoretical impact} of the granularity of preference supervisions on model stability.
Our theorem suggests that training with scaled supervision via more fine-grained preferences can be formulated as an implicit Lipschitz regularizer for CTR models to achieve better stability and generalizability.
In the context of CTR prediction, we propose replacing binary labels with fine-grained preference feedback to guide the training of a CTR model.
With more scaled supervision,
the model's output becomes smoother and more consistent, resulting in less overfitting short-term fluctuations hidden in the data with improved model generalization and stability in learning high-level interaction patterns.

In this work, we leverage the continuous nature of ratings in user feedback to increase the bandwidth of supervision signals beyond current Boolean labels in CTR prediction with explicit feedback.
By modeling ratings categorically, we enable the learning of high-level interaction patterns across distinct rating levels in a more segmented and independent manner. 
These categorical ratings are subsequently transformed into binary labels, aligning with the original CTR prediction task while preserving the enriched supervision benefits.
Our contributions can be summarized as follows:
\begin{itemize}[leftmargin=*]
    \item We theoretically identify the impact of the granularity of preference supervisions on a deep model's stability: more refined supervisions help reduce a model's Lipschitz constant, resulting in more stable and smoother model output.
    \item Inspired by the theorem, we identify the limitations of binary labeling in capturing the full spectrum of user preferences in modern social media recommender systems, and then propose to use more fine-grained preferences to increase the bandwidth of supervisions for more enhanced model stability and generalizability.
    \item Extensive experiments based on our open-release implementation and datasets are conducted to demonstrate that our approach significantly improves CTR prediction performance and promotes better learning dynamics in optimizing model parameters.
\end{itemize}

\section*{Preliminary}
\subsection*{Lipschitz Constant in Deep Models}
A function $f:\mathbb{R}^{n} \rightarrow \mathbb{R}^{m}$ is said to be Lipschitz continuous on an input set $\mathcal{X}\subseteq  \mathbb{R}^{n}$ if there exists a bound $K \geq 0$ such that for all $\vx,\vy \in \mathcal{X}$, $f$ satisfies:
\begin{equation}
\label{eq-1}
    \norm{{f(\vx) - f(\vy)}} \le K\norm{\vx-\vy}, \quad \forall \vx,\vy \in \mathcal{X}.
\end{equation}
The smallest possible $K$ in Equation~(\ref{eq-1}) is defined as the Lipschitz constant of $f$, denoted as ${\rm Lip}(f)$:
\begin{equation}
\label{eq-2}
    {\rm Lip}(f) = \sup\limits_{\vx,\vy\in \mathcal{X}, \vx\ne \vy}\frac{\norm{f(\vx)-f(\vy)}}{\norm{\vx-\vy}}.
\end{equation}
In this context, $f$ is known as a $K$-Lipschitz function, where the Lipschitz constant quantifies the maximum change in the function’s output resulting from a unit-norm perturbation in its input. 
This constant is a crucial indicator of a deep model’s stability wrt the inputs. 
However, determining the exact Lipschitz constant is computationally challenging. 
As highlighted in~\cite{virmaux2018lipschitz}, computing the exact Lipschitz constant for deep models has been proven to be NP-hard.
\begin{figure*}[t]
\centering
\includegraphics[width=1\textwidth]{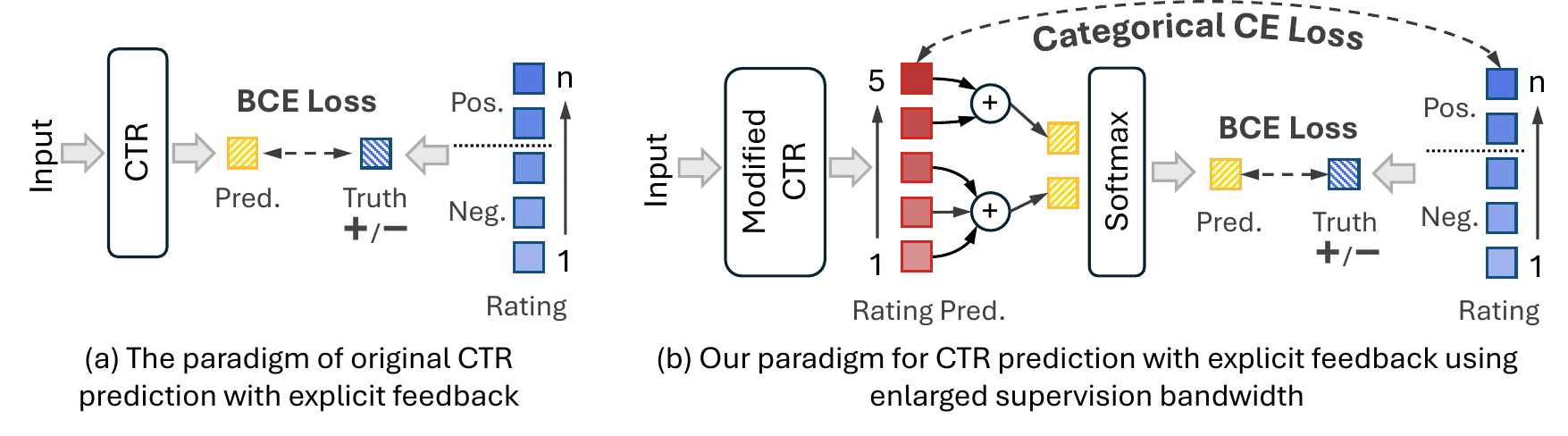}
\caption{(a) and (b) demonstrate the existing and our modified paradigm of training CTR models with explicit feedback respectively. The input includes user, item IDs and the contextual information of the interaction in between. Our paradigm enlarges the supervision bandwidth from explicit preferences: we slightly modify existing CTR models' prediction layers to recover the explicit preferences for more refined supervision. These logits are then normalized via the Softmax function to the probability that matches the CTR prediction task. }
\label{fig:model}
\end{figure*}

\subsection*{CTR Prediction in Modern RecSys}
\label{sec:ctr}
\paragraph{General CTR Prediction Paradigm}
Click-through rate (CTR) prediction is a task focused on estimating the probability that a user will engage with a specific item. 
This prediction takes the user and item IDs, features related to the user, the item, and the broader context in which the interaction occurs as the input, and the expected output is a probability likelihood ranging from zero to one.
Formally, we denote the IDs of user $i$ and item $j$ as $x_i$ and $x_j$, respectively.
We denote the raw contextual features (including user features, item features, and features related to the interaction context such as duration time) of the interaction in between as $\textbf{c}_{ij} \in \mathbb{R}^{d^c}$, where $d^c$ refers to the dimension of the contextual feature.
The user ID $x_i$, item ID $x_j$, and the raw contextual features $\textbf{c}_{ij}$ construct the input of the CTR prediction.
The input is then encoded through the embedding layers to be transformed into dense vectors in a high-dimensional space.

Formally, we encode user/item IDs with the function $f(\cdot):\mathbb{R} \rightarrow \mathbb{R}^d$, where $d$ refers to the latent dimension of the encoded ID embeddings.
Similarly, we encode the contextual features with the function $h(\cdot): \mathbb{R}^{d^c} \rightarrow \mathbb{R}^{d'}$, where $d'$ is the dimension of the encoded contextual embeddings.
The encoded input $\textbf{z}_{ij} \in \mathbb{R}^{2d+d'}$ is defined as the concatenation of the three encoded features.
The above embedding process can be summarized as:
\begin{align}
    \textbf{z}_i & =  f(x_i), \; \textbf{z}_j = f(x_j), \; \textbf{z}^c_{ij} = h(\textbf{c}_{ij}), \label{eq:enc_1} \\
    \textbf{z}_{ij} & = \; \left[\textbf{z}_i \; \| \; \textbf{z}_j \; \| \; \textbf{z}^c_{ij}\right],
    \label{eq:enc_2}
\end{align}
where $\|$ refers to the concatenation operation, $\textbf{z}_i$ and $\textbf{z}_j$ refer to the encoded ID embeddings of user $i$ and item $j$ respectively, and $\textbf{z}^c_{ij}$ are the encoded contextual features. 
After the embedding process, we feed the embeddings to the learnable operations to learn both low-level and high-level interaction patterns.
Specifically, we denote the learnable operations as $q(\cdot): \mathbb{R}^{2d+d'} \rightarrow \mathbb{R}$, and let $p_{ij} = q(\textbf{z}_{ij})$, where $p_{ij} \in [0, 1]$ represents how likely user $i$ would interact with item $j$.
We depict the CTR prediction paradigm in Figure~\ref{fig:model} (a).

\paragraph{DCN as an Examplification}
The learnable operations in different CTR models vary by their intrinsic model structures.
To illustrate the learning scheme of CTR prediction, we here demonstrate the modeling process of DCN~\cite{wang2021dcn}, one of the most typical CTR models commonly used in both industrial and academic environments. 
There are two distinct types of trainable layers in DCN: the cross layers and the deep layers, which constitute the cross and the deep network respectively.
The cross-network is designed to model explicit feature interactions. 
Each cross-layer is formulated as
\begin{equation}
    \textbf{z}^{(l+1)} = \textbf{z}^{(0)}{\textbf{z}^{(l)}}^\intercal \textbf{w}^{(l)} + \textbf{b}^{(l)} + \textbf{z}^{(l)},
\end{equation}
where $\textbf{z}^{(l+1)}, \textbf{z}^{(l)} \in \mathbb{R}^d$ are the input and output column vectors from the $l$-th cross layer, and $\textbf{w}^{(l)}, \textbf{b}^{(l)} \in \mathbb{R}^d$ are the trainable parameters in the $l$-th cross layer.
The deep network is a series of fully connected layers designed to capture complex and nonlinear interactions:
\begin{equation}
    \textbf{h}^{(l+1)} = \text{ReLU}(\textbf{w}^{(l+1)}\textbf{h}^{(l)} + \textbf{b}^{(l)}),
\end{equation}
where $\textbf{h}^{(l)} \in \mathbb{R}^{d_l}, \textbf{h}^{(l+1)} \in \mathbb{R}^{d_{l+1}}$ are the input and output of the $l$-th hidden layer, respectively, and $\textbf{w}^{(l)} \in \mathbb{R}^{d_{l+1} \times d_l}$ are the trainable parameters in the $l$-th layer.

For a given user $i$ and item $j$, we let the input of the first layers as the encoded features. 
That is, $\textbf{z}^{(0)} = \textbf{h}^{(0)} = \textbf{z}_{ij}$.
The output of both networks are then concatenated and fed into a prediction layer to generate the final binary prediction:
\begin{equation}
\label{eq:dcn-logits}
    p_{ij} = \sigma \left( [\textbf{z}^{(L_1)} \; \| \; \textbf{h}^{(L_2)} ] \textbf{w}_{\text{logits}} \right),
\end{equation}
where $\textbf{z}^{(L_1)} \in \mathbb{R}^{d_1}, \textbf{h}^{(L_2)} \in \mathbb{R}^{d_2}$ are the $L_1$-th and $L_2$-th layer outputs from the cross and deep networks, respectively, $\textbf{w}_{\text{logits}} \in \mathbb{R}^{d_1 + d_2}$ is the weight vector in the logits layer, and $\sigma(x) = 1 / (1 + e^{-x})$.
For the binary classification task formalized for the CTR prediction task, the DCN model is trained with the binary cross entropy (BCE) loss:
\begin{align}
\label{eq:dcn-loss}
    \mathcal{L} = & -\frac{1}{N} \sum_{\{i,j\}\in T_r} y_{ij} \log(p_{ij}) + (1-y_{ij}) \log(1-p_{ij}) \notag \\
    & + \lambda \sum_l \Vert\textbf{w}^{(l)} \Vert ^ 2,
\end{align}
where $T_r$ denotes the training set of positive and negative pairs, $p_{ij}$ is the predicted interaction probability between user $i$ and item $j$, $y_{ij}$ refers to the binary label for the user-item engagement (1 for positive and 0 for negative pairs), $\lambda$ is the $L_2$ regularization coefficient, and $\textbf{w}^{(l)}$ represents the parameters of the CTR model.

\section*{Estimating Lipschitz Effects of Preference Feedback}
We first conduct a theoretical analysis of how the granularity of preference feedback influences the Lipschitz constant of deep models.
Then, in light of our theoretical analysis, 
instead of seeking for a proxy to approximate the Lipschitz constant,
we propose a strategy that utilizes fine-grained ratings to enhance existing CTR methods' stability and generality.
Our strategy acts as an \textit{implicit} Lipschitz regularizer and can be naturally derived without complex math derivations.
It simply encourages the models to generate smoother output to mitigate the overfitting problem by capturing more general user-item interactive patterns.

\subsection*{More Fine-grained Supervision Enhances \\Model Stability}

In this section, we present a theoretical foundation to explore the relationship between supervision bandwidth and model stability. 
Specifically, we refer the dimension of the output logits $N$ as the supervision bandwidth and analyze how $N$ influences the Lipschitz continuity of the model's output.

\begin{tcolorbox}[width=1.0\linewidth, colback=citecolor!5, colframe=black, arc=0pt, boxsep=0mm, arc=2mm, left=2mm, right=2mm, top=5mm, bottom=2mm]
\vspace{-0.3cm}
\begin{theorem}
\label{thm-1}
{Let $\mathbf{f}(\mathbf{x}) = [f_1(\mathbf{x}), \dots, f_N(\mathbf{x})]$ be the logits output by a model for an input $\mathbf{x} \in \mathcal{X} \subset \mathbb{R}^D$, where $N$ is the dimension of the output logits. Let $\mathbf{p}(\mathbf{x}) = \sigma(\mathbf{f}(\mathbf{x}))$ denote the corresponding normalized probabilities with temperature scaling. Assume that the model's logits $\mathbf{f}(\mathbf{x})$ are Lipschitz continuous with respect to $\mathbf{x}$ with Lipschitz constant $L_f$. Then the Lipschitz constant $L_p(N)$ of the normalization function wrt the input $\mathbf{x}$ satisfies:}
\begin{equation}
\label{eq:thm-1}
L_p(N) \leq \frac{L_f}{\sqrt{N}}.
\end{equation}
\end{theorem}
\end{tcolorbox}

\noindent
\textit{Insight.} Theorem~\ref{thm-1} indicates that as the supervision bandwidth $N$ increases, the Lipschitz constant of the normalization function decreases.
A decreased Lipschitz constant implies that the model is less prone to change drastically in response to minor variations in the input, thereby reflecting better model stability and generality.

\begin{proof}[Proof Sketch]
Let $\mathbf{f}(\mathbf{x}) = [f_1(\mathbf{x}), \dots, f_N(\mathbf{x})]$ denote the logits output by the model for input $\mathbf{x}$, where $N$ is the number of logits. The softmax normalization function with temperature scaling is given by:
\begin{equation}
p_i(\mathbf{x}) = \frac{\exp\left(f_i(\mathbf{x})/\tau\right)}{\sum_{j=1}^{N} \exp\left(f_j(\mathbf{x})/\tau\right)} \quad \text{for } i = 1, \dots, N,
\end{equation}
where $\tau > 0$ adjusts the uniformity of the distribution.
To analyze the Lipschitz constant $L_p(N)$ of the model’s output probabilities, we consider the Frobenius norm of the Jacobian matrix $\mathbf{J}_{\sigma}(\mathbf{f})$:
\begin{equation}
\frac{\partial p_i}{\partial f_j} = p_i(\delta_{ij} - p_j),
\end{equation}
where $\delta_{ij}$ is the Kronecker delta. The norm quantifies the sensitivity of the output probabilities to changes in the logits.
Assuming evenly distributed logits as \( N \) increases, $p_i \approx \frac{1}{N}$ for all $i$ \cite{muller2019does}, and the Jacobian elements are approximated as:
\begin{equation}
\frac{\partial p_i}{\partial f_j} \approx \frac{1}{N} (\delta_{ij} - \frac{1}{N}).
\end{equation}
The Frobenius norm of the Jacobian simplifies to:
\begin{equation}
\|\mathbf{J}_{\sigma}(\mathbf{f})\|_F \approx \frac{1}{\sqrt{N}}.
\end{equation}
Given that the logits $\mathbf{f}(\mathbf{x})$ are Lipschitz continuous with constant $L_f$, the sensitivity of the probabilities with respect to the input $\mathbf{x}$ is bounded as:
\begin{equation}
\|\mathbf{p}(\mathbf{x}) - \mathbf{p}(\mathbf{y})\| \leq \|\mathbf{J}_{\sigma}(\mathbf{f})\|_F \cdot L_f \|\mathbf{x} - \mathbf{y}\|.
\end{equation}
Thus, the Lipschitz constant $L_p(N)$ satisfies:
\begin{equation}
L_p(N) \leq \frac{L_f}{\sqrt{N}},
\end{equation}
where the reduced Frobenius norm ensures that as $N$ increases, the model becomes less sensitive to input variations \cite{miyato2018spectral}.
\end{proof}

The proof sketch provides the following key insight: as the supervision bandwidth $N$ increases, the Lipschitz constant of the model under this finer-grained supervision decreases.
In other words, as the preference feedback become more fine-grained, the norm of the Jacobian matrix is reduced, resulting in a smoother gradient landscape.
Correspondingly, this reduction in the Lipschitz constant, as reflected in model gradients, decreases the sensitivity of output probabilities to input variations.
This smoothing effect effectively acts as an implicit form of gradient regularization, thereby enhancing model stability~\cite{neyshabur2018the}.
Importantly, the benefits of this regularization grow as the complexity of the classification task increases with larger supervision bandwidth (i.e., as $N$ becomes larger).

\subsection*{Stabilizing CTR Models via Fine-grained Preference Supervision}
Following our prior study, to overcome the limitations of current CTR models which rely on binary labels to provide training supervision, we propose a strategy that exploits the fine-grained nature in user feedback ratings to provide more detailed supervision signals. 
In the real world, explicit preference feedback is expressed through ratings $\textbf{r}_{ij}$, where $r_{ij} \in \{1, 2, 3, ..., N\}$, reflecting varying degrees of interest and satisfaction with item $j$ by user $i$. 
Traditional CTR models convert these ratings into binary labels $y_{ij} \in \{0, 1\}$ based on thresholding.
Interactions with ratings higher than the threshold are labeled as positive ones, and those lower than the threshold are labeled as negative ones.
In our approach, we additionally adopt the fine-grained ratings as the ground truth for supervision.
As shown in Figure~\ref{fig:model} (b), our approach modifies the traditional CTR model to be supervised by the discrete ratings directly. 

To adapt existing CTR models to handle these fine-grained ratings, we re-formulate the problem as a multi-class classification problem and modify their prediction layers to output explicit rating predictions. 
Let the modified rating prediction function be $q'(\cdot): \mathbb{R}^{2d+d'} \rightarrow \mathbb{R}$, $\hat{r}_{ij} = q'(\textbf{z}_{ij})$, where $\hat{r}_{ij}$ represents the predicted rating for user $i$ and item $j$. 
The predicted ratings are then supervised with the categorical Cross Entropy (CE) loss, defined as:
\begin{equation}
    \mathcal{L}_{\text{CE}} = -\frac{1}{N} \sum_{\{i,j\}\in T_r} \sum_{k=1}^{K} y_{ij}^k \log p_{ij}^k,
\end{equation}
where $T_r$ denotes the training set with pairs of user $i$ and item $j$ and their labels, $y_{ij}^k$ is the true rating label in one-hot encoding, and $p_{ij}^k$ is the predicted probability for rating $k$.
Note that adopting CE loss plays a crucial role in our method by heavily penalizing over-confident incorrect predictions through large logarithmic losses.
This mechanism discourages skewed or overly confident logits, promoting an even distribution of logits. 
This aligns with the assumption of evenly distributed logits, which underpins the proof of Theorem~\ref{thm-1}.
To ensure compatibility with traditional CTR tasks, the explicit rating predictions are then aggregated into a probability that matches the binary prediction. 
Let $\hat{p}_{ij}$ be the aggregated probability that user $i$ will click on item $j$, defined as
\begin{equation}
    \hat{p}_{ij} = \frac{\sum_{k > t_{sh}}{p_{ij}^k}}{\sum_{k\leq t_{sh}}{p_{ij}^k} + \sum_{k > t_{sh}}{p_{ij}^k}},
\end{equation}
where $p_{ij}^k$ is the predicted probability for rating $k$ and $t_{sh}$ is the rating threshold defining the positive and negative feedback.
This aggregated probability $\hat{p}_{ij}$ can then be used in the traditional BCE loss for final supervision:
\begin{align}
    \mathcal{L}_{\text{BCE}} = & -\frac{1}{N} \sum_{\{i,j\}\in T_r} y_{ij} \log(\hat{p}_{ij}) + (1-y_{ij}) \log(1-\hat{p}_{ij}) \notag \\
    & + \lambda \sum_l \Vert\hat{\textbf{w}}^{(l)} \Vert ^ 2,
\end{align}
where $y_{ij}$ is the binary label indicating whether user $i$ clicked on item $j$, $\hat{\textbf{w}}^{(l)}$ represents the parameters of the modified CTR model.
The final supervised loss $\mathcal{L}$ is defined as:
\begin{equation}
    \mathcal{L} = \lambda_r \mathcal{L}_{CE} + (1 - \lambda_r) \mathcal{L}_{BCE},
\end{equation}
where $\lambda_r \in [0, 1]$ is the contributing ratio of the categorical cross-entropy loss.

\subsection*{Boosting user engagement from the perspective of behavioral psychology}
Prior research in behavioral psychology examines the relationship between user engagement and recommender systems~\cite{pachali2024drives,zhang2024keep}, showing that platform-driven features (e.g., playlists on Spotify) significantly boost interaction.
Additionally, factors like user satisfaction and dependence play crucial roles in sustaining engagement. 
Our approach, which stabilizes CTR models for improved generality, aligns with these findings enabling the models to better capture and respond to user preferences. 
This is accomplished by focusing on learning general user behaviors rather than overfitting to noisy interactions, such as bait clicks or non-engaged interactions.
From the end user’s perspective, our method results in more personalized and accurate recommendations, reducing exposure to irrelevant content and providing a more satisfying and tailored experience. 
Enhanced user satisfaction and engagement not only improve the user experience but also deliver substantial commercial and psychological benefits, creating a mutually beneficial outcome for both users and companies. 
We show a concrete example for end user experience improvement in recommendation in Appendix~\ref{appd:end_user}.

\section*{Experiments}
We evaluate the predictive and ranking performance of applying our strategy to multiple popular baselines used in social media RecSys. To assess stability, we compare the gradient norms before and after applying our approach to a representative CTR model. Additionally, we demonstrate that our method consistently outperforms popular baselines, even without exhaustive hyperparameter tuning. Lastly, we report the average computational time before and after integrating our approach into the baselines to demonstrate its superior practicability to real-world applications.

\subsection*{Setup} \label{sec:exp-setup}
\subsubsection*{Datasets}
We select three publicly available and popular recommendation benchmark datasets to conduct the experiments:
\textit{(1)} \textbf{MovieLens-1M}\footnote{https://grouplens.org/datasets/movielens/1m/} dataset is sourced from the MovieLens website, containing 1 million ratings from 6,000 users on 4,000 movies;
\textit{(2)} \textbf{Yelp2018}\footnote{https://www.kaggle.com/datasets/yelp-dataset/yelp-dataset/} dataset includes a large collection of user reviews, item information, and user-item interactions from the Yelp platform, with millions of ratings, reviews, and other data points such as item categories and user profiles;
\textit{(3)} \textbf{Amazon-Book}\footnote{http://jmcauley.ucsd.edu/data/amazon/} dataset is derived from the Amazon product data and contains detailed information on user interactions with books, including millions of ratings, reviews, and metadata such as book categories, titles, and author information.
Based on data sparsity, we categorize ML-1M as a dense dataset, and Yelp2018 and Amazon-Book are identified as sparse datasets.
The dataset statistics are provided in Appendix~\ref{appd:data_stats}.
The feedback provided by users in the above datasets is explicit, with the format of ratings from 1 to 5.
For all the feedback, we convert the ratings to binary labels through thresholding and set the threshold as 3.
We randomly split datasets with a ratio of 0.8/0.1/0.1 for training, validation, and testing, respectively.

\subsubsection{Baselines}
We select seven CTR baseline models that specifically focus on feature interaction modeling, including WideDeep~\cite{cheng2016wide}, NFM~\cite{he2017neural}, xDeepFM~\cite{lian2018xdeepfm}, AutoInt~\cite{song2019autoint}, FiGNN~\cite{li2019fi}, DCNV2~\cite{wang2021dcn}, and EulerNet~\cite{tian2023eulernet}.
Focusing on models designed specifically for feature interaction modeling isolates the impact of supervision bandwidth enlargement effectively. 
Including diverse model types with differing objectives (e.g., behavior prediction, auxiliary tasks) could introduce modeling biases, confounding the true effects of our supervision bandwidth modification.
Specifically:
NFM~\cite{he2017neural} and xDeepFM~\cite{lian2018xdeepfm} combine the advantages of Factorization Machines~\cite{rendle2010factorization} and deep neural networks to capture complex non-linear and high-order feature interactions.
DCNV2~\cite{wang2021dcn} learns explicit and implicit feature interactions through a cross-network and a deep neural network, respectively. It improves DCN with a low-rank cross-network that enhances the efficiency and interpretability of the model.
AutoInt~\cite{song2019autoint} utilizes self-attentive neural networks to learn more effective feature interactions.
EulerNet~\cite{tian2023eulernet} learns high-order feature interactions by transforming their exponential powers into linear combinations of the modulus and phase of complex features.

\subsubsection{Training configurations}
For all the baselines, we employ the AdamW~\cite{loshchilov2018decoupled} optimizer for optimization.
We run a fixed number of grid searches over all the baseline models' provided hyper-parameters for their best AUC performance on the validation set.
With consistent hyper-parameters, we train the models under five random seeds and save all the checkpoints.
We repeat the hyper-parameter search process for our method as well.
The models' CTR prediction abilities are mainly evaluated by two predictive metrics, AUC and logloss.
A higher AUC indicates that the model more accurately distinguishes whether an item is of interest to the user or not.
In addition, we evaluate the ranking abilities of the CTR models, i.e., how well the models rank the more relative/highly rated items.
We do so to emphasize that the recommendation task is essentially a ranking task, therefore it is crucial to preserve the ranking abilities of the CTR models.
For the ranking metrics, we select NDCG@K and recall@k, and set $k$ to 10 and 20.
We adopt the recommender system library named RecBole~\cite{zhao2021recbole} to conduct all the experiments.

\begin{table}[t]
\caption{The AUC and logloss of the CTR models before and after applying our strategy. The \textcolor[HTML]{4169E1}{\textbf{Original}} column represents the original CTR model, the \textcolor[HTML]{B22222}{\textbf{Smooth}} column denotes the results after applying our approach, and the \textbf{$\Delta\%$} column shows the relative change of the performance in terms of the original models' performance.}
\label{tab:main_exp}
\centering
\begin{adjustbox}{width=\linewidth,center}
\begin{tabular}{c!{\vrule width \lightrulewidth}ccc!{\vrule width \lightrulewidth}ccc} 
\toprule
\multicolumn{7}{c}{{(i) ML-1M}}                                                                                                      \\ 
\midrule
\textbf{Metric} & \multicolumn{3}{c!{\vrule width \lightrulewidth}}{\textbf{AUC(\%)} $\uparrow$} & \multicolumn{3}{c}{\textbf{LogLoss ($\times 100$)} $\downarrow$}                \\ 
\midrule
\textbf{Model}  & \textcolor[HTML]{4169E1}{\textbf{Original}}& \textcolor[HTML]{B22222}{\textbf{Smooth}} & \textbf{$\uparrow\Delta\%$}              & \textcolor[HTML]{4169E1}{\textbf{Original}}& \textcolor[HTML]{B22222}{\textbf{Smooth}} & \textbf{$\downarrow\Delta\%$}  \\ 
\midrule
WideDeep        & 82.10             & 82.46           & 0.44                      & 50.83             & 50.50           & -0.65         \\
NFM             & 81.82             & 82.02           & 0.25                      & 51.28             & 51.30           & 0.04          \\
xDeepFM         & 81.38             & 82.43           & 1.29                      & 52.07             & 51.14           & -1.79         \\
AutoInt         & 82.09             & 82.56           & 0.58                      & 51.05             & 50.36           & -1.34         \\
FiGNN           & 81.58             & 81.66           & 0.10                      & 52.04             & 51.90           & -0.27         \\
DCNV2           & 82.16             & 82.61           & 0.55                      & 51.04             & 50.43           & -1.19         \\
EulerNet        & 82.03             & 82.00           & -0.04                     & 51.39             & 51.62           & 0.45          \\ 
\midrule
\multicolumn{7}{c}{{(ii) Yelp2018}}                                                                                                   \\ 
\midrule
WideDeep        & 74.59             & 74.96           & 0.49                      & 55.20             & 55.02           & -0.32         \\
NFM             & 74.12             & 74.55           & 0.57                      & 55.56             & 55.23           & -0.60         \\
xDeepFM         & 74.55             & 74.89           & 0.46                      & 55.28             & 55.03           & -0.46         \\
AutoInt         & 74.57             & 74.99           & 0.56                      & 55.27             & 55.04           & -0.42         \\
FiGNN           & 74.59             & 74.89           & 0.40                      & 55.28             & 55.36           & 0.14          \\
DCNV2           & 74.43             & 74.78           & 0.47                      & 55.32             & 55.18           & -0.25         \\
EulerNet        & 74.54             & 74.78           & 0.32                      & 55.21             & 55.04           & -0.31         \\ 
\midrule
\multicolumn{7}{c}{{(iii) Amazon-book}}                                                                                                \\ 
\midrule
WideDeep        & 80.95             & 81.42           & 0.58                      & 39.01             & 38.57           & -1.11         \\
NFM             & 80.73             & 81.22           & 0.60                      & 39.23             & 38.92           & -0.79         \\
xDeepFM         & 80.95             & 81.39           & 0.54                      & 39.04             & 39.76           & 1.85          \\
AutoInt         & 80.97             & 81.37           & 0.50                      & 38.97             & 38.70           & -0.70         \\
FiGNN           & 80.96             & 81.25           & 0.36                      & 39.07             & 39.03           & -0.08         \\
DCNV2           & 80.89             & 81.27           & 0.47                      & 39.01             & 38.87           & -0.36         \\
EulerNet        & 80.97             & 81.26           & 0.36                      & 38.97             & 39.28           & 0.79          \\
\bottomrule
\end{tabular}
\end{adjustbox}
\end{table}

\begin{table*}[ht]
\caption{The ranking results of the CTR models \textit{before and after} applying our implicit Lipschitz regularizer. The \textcolor[HTML]{4169E1}{\textbf{Original}} column represents the original CTR model, the \textcolor[HTML]{B22222}{\textbf{Smooth}} column denotes the results after applying our approach, and the \textbf{$\Delta\%$} column shows the relative change of the performance in terms of the original models' performance.}
\label{tab:ranking}
\centering
\begin{adjustbox}{width=\linewidth,center}
\begin{tabular}{c!{\vrule width \lightrulewidth}ccc!{\vrule width \lightrulewidth}ccc!{\vrule width \lightrulewidth}ccc!{\vrule width \lightrulewidth}ccc} 
\toprule
\textbf{Metric} & \multicolumn{3}{c!{\vrule width \lightrulewidth}}{\textbf{NDCG@10$~\uparrow$}} & \multicolumn{3}{c!{\vrule width \lightrulewidth}}{\textbf{NDCG@20$~\uparrow$}} & \multicolumn{3}{c!{\vrule width \lightrulewidth}}{\textbf{Recall@10$~\uparrow$}} & \multicolumn{3}{c}{\textbf{Recall@20$~\uparrow$}}              \\ 
\midrule
\textbf{Model}  & \textcolor[HTML]{4169E1}{\textbf{Original}}& \textcolor[HTML]{B22222}{\textbf{Smooth}} & \textbf{$\uparrow\Delta\%$}                  & \textcolor[HTML]{4169E1}{\textbf{Original}}& \textcolor[HTML]{B22222}{\textbf{Smooth}} & \textbf{$\uparrow\Delta\%$}                  & \textcolor[HTML]{4169E1}{\textbf{Original}}& \textcolor[HTML]{B22222}{\textbf{Smooth}} & \textbf{$\uparrow\Delta\%$}                    & \textcolor[HTML]{4169E1}{\textbf{Original}}& \textcolor[HTML]{B22222}{\textbf{Smooth}} & \textbf{$\uparrow\Delta\%$}  \\ 
\midrule
\multicolumn{13}{c}{(i) ML-1M}                                                                                                                                                                                                                                                                \\ 
\midrule
WideDeep        & 11.70             & 11.72           & 0.21                          & 14.12             & 14.34           & 1.56                          & 13.49             & 13.86           & 2.73                            & 21.17             & 21.92           & 3.51          \\
NFM             & 11.90             & 12.34           & 3.72                          & 14.57             & 15.11           & 3.66                          & 14.21             & 15.04           & 5.85                            & 22.60             & 23.55           & 4.21          \\
xDeepFM         & 11.97             & 13.65           & 14.09                         & 14.62             & 16.24           & 11.09                         & 14.31             & 15.67           & 9.52                            & 22.63             & 24.11           & 6.53          \\
AutoInt         & 11.16             & 12.24           & 9.73                          & 13.68             & 14.78           & 8.10                          & 13.06             & 14.08           & 7.78                            & 20.91             & 22.09           & 5.65          \\
FiGNN           & 11.48             & 12.19           & 6.17                          & 14.21             & 14.86           & 4.53                          & 14.10             & 14.66           & 3.96                            & 22.45             & 22.93           & 2.12          \\
DCNV2           & 11.54             & 13.16           & 14.07                         & 14.01             & 15.66           & 11.79                         & 13.52             & 14.93           & 10.45                           & 21.21             & 22.97           & 8.32          \\
EulerNet        & 11.58             & 12.58           & 8.69                          & 14.33             & 15.17           & 5.82                          & 14.11             & 14.79           & 4.88                            & 22.46             & 22.96           & 2.20          \\ 
\midrule
\multicolumn{13}{c}{(ii) Yelp2018}                                                                                                                                                                                                                                                             \\ 
\midrule
WideDeep        & 3.05              & 3.14            & 3.22                          & 5.29              & 5.38            & 1.66                          & 6.15              & 6.34            & 3.19                            & 14.52             & 14.66           & 0.92          \\
NFM             & 3.78              & 3.97            & 4.92                          & 6.16              & 6.38            & 3.70                          & 7.57              & 7.89            & 4.23                            & 16.29             & 16.78           & 2.98          \\
xDeepFM         & 3.36              & 3.05            & -9.11                         & 5.65              & 5.27            & -6.80                         & 6.63              & 6.21            & -6.28                           & 15.11             & 14.45           & -4.35         \\
AutoInt         & 2.99              & 3.22            & 7.56                          & 5.28              & 5.52            & 4.62                          & 6.15              & 6.55            & 6.40                            & 14.67             & 15.12           & 3.03          \\
FiGNN           & 3.09              & 3.16            & 2.26                          & 5.37              & 5.49            & 2.16                          & 6.28              & 6.48            & 3.09                            & 14.76             & 15.12           & 2.45          \\
DCNV2           & 2.99              & 2.95            & -1.20                         & 5.35              & 5.25            & -1.87                         & 6.29              & 6.21            & -1.34                           & 15.07             & 14.75           & -2.11         \\
EulerNet        & 4.44              & 4.33            & -2.57                         & 6.73              & 6.67            & -0.98                         & 8.11              & 8.01            & -1.21                           & 16.55             & 16.64           & 0.51          \\ 
\midrule
\multicolumn{13}{c}{(iii) Amazon-book}                                                                                                                                                                                                                                                          \\ 
\midrule
WideDeep        & 2.42              & 2.71            & 12.24                         & 3.76              & 4.16            & 10.53                         & 4.08              & 4.69            & 15.11                           & 8.69              & 9.57            & 10.06         \\
NFM             & 2.49              & 2.47            & -0.88                         & 3.82              & 3.89            & 1.88                          & 4.19              & 4.37            & 4.25                            & 8.73              & 9.20            & 5.38          \\
xDeepFM         & 3.36              & 3.05            & -9.11                         & 3.60              & 3.60            & 0.00                          & 3.84              & 4.05            & 5.25                            & 8.31              & 8.74            & 5.20          \\
AutoInt         & 2.38              & 2.68            & 12.77                         & 3.76              & 4.13            & 9.73                          & 4.07              & 4.63            & 13.75                           & 8.83              & 9.51            & 7.70          \\
FiGNN           & 2.38              & 2.54            & 6.72                          & 3.72              & 3.96            & 6.45                          & 3.97              & 4.43            & 11.54                           & 8.58              & 9.23            & 7.65          \\
DCNV2           & 2.83              & 2.89            & 2.19                          & 4.53              & 4.64            & 2.61                          & 5.10              & 5.40            & 5.88                            & 11.02             & 11.44           & 3.79          \\
EulerNet        & 2.38              & 3.05            & 28.24                         & 3.76              & 4.46            & 18.66                         & 4.07              & 4.97            & 21.94                           & 8.83              & 9.71            & 10.04         \\
\bottomrule
\end{tabular}
\end{adjustbox}
\end{table*}

\begin{figure*}[ht]
  \centering
  \includegraphics[width=\textwidth]{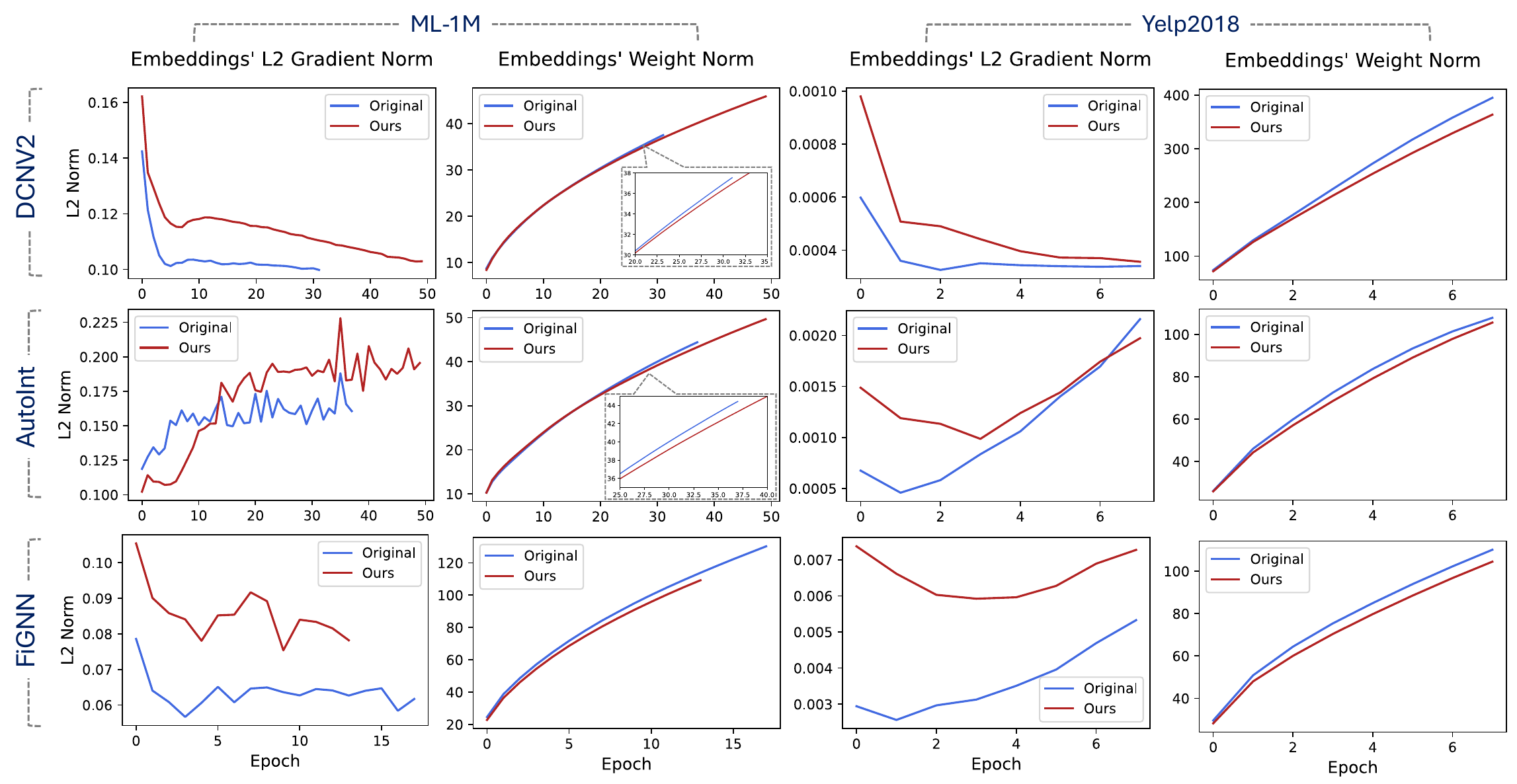}
  \caption{The L2 gradient norms during training and the weight norms post-training for the learned user and item embeddings in the \textcolor[HTML]{4169E1}{\textbf{original}} DCNV2, AutoInt, and FiGNN methods, and those after applied with \textcolor[HTML]{B22222}{\textbf{our}} approach.}
  \label{fig:grad_norm}
\end{figure*}

\subsection*{Improved predictive and ranking performance} \label{sec:exp-main}
\paragraph{Predictive Improvement}
For all the baseline models, we first evaluate their recommendation performance on the three benchmark datasets by the predictive metrics, AUC (the higher the better) and logloss (the lower the better).
The results are shown in Table~\ref{tab:main_exp}, where the \textcolor[HTML]{4169E1}{\textbf{Original}} column shows the performance of the original CTR models, the \textcolor[HTML]{B22222}{\textbf{Smooth}} column shows the performance of applying our method to the corresponding CTR models, and \textbf{$\Delta\%$} represents the relative improvement wrt the original performance.
From the table, we observe that: 
(i) while the degree of improvement varies across the combination of CTR models and benchmark datasets, our strategy stably improves the AUC over the original model.
On average, we observe $\sim$0.47\% AUC improvement.
The improvement is particularly significant for the CTR prediction task compared to prior methods focused on enhancing feature interaction modeling. For instance, in terms of AUC on the Movielens dataset, DCNV2~\cite{wang2021dcn}, AFN~\cite{cheng2020adaptive}, and FinalNet~\cite{zhu2023final} achieve relative gains of approximately 0.1\%, 0.05\%, and 0.07\% over the previous state-of-the-art, respectively.
The improved AUC results validate that our strategy is effective in improving the CTR model's ability to accurately identify items of interest, across various model structures and benchmark datasets.
(ii) the values of logloss are generally decreased, suggesting more stable and confident predictions.
In some rare cases, we observe slight increments in the logloss.
However, this slight increase is acceptable with the corresponding improved AUC, which is directly associated with the quality of the recommendation.

\paragraph{Ranking Improvement}
In addition to the predictive metrics, we also assess the impact of our strategy on the CTR models' ranking capabilities, which is ranking more relevant items higher than the less relevant ones. 
Although predictive metrics such as AUC and logloss are well-aligned with the CTR prediction task, they do not fully capture the practical nuances in real-world recommendations. 
One thing to constantly bear in mind is that recommendation is fundamentally a ranking task. 
In practice, the predictive probabilities of a CTR model are commonly used as the ranking score to rank recommended items.
Thus, while enhancing the predictive ability of a CTR model, it is equally important to ensure that its ranking performance is well-preserved.
We compare the ranking metrics of CTR models before and after applying our approach and demonstrate the results in Table~\ref{tab:ranking}.

From Table~\ref{tab:ranking}, we observe that our strategy not only maintains the ranking capabilities of the original CTR models, but also improves them. While there are a few instances of performance degradation, the overall improvements are remarkable. Our strategy acts as a regularizer, resulting in a more balanced distribution of probabilities across rating classes. This smoother prediction reduces the risk of overconfidence by preventing excessively high ranking scores for certain user-item pairs, thus improving the model's ability to distinguish between relevant and irrelevant items and leading to more refined ranking lists. Although the ranking metrics are not directly optimized by our auxiliary objective (i.e., categorical loss), the Lipschitz regularization statistically improves ranking performance. This improvement is significant even for the Yelp2018 dataset, which includes a wide variety of businesses such as restaurants, hotels, shopping malls, and services. The broader scope of Yelp2018 introduces more variability and complexity compared to the other two datasets.

\begin{figure*}[ht]
  \centering
  \includegraphics[width=\textwidth]{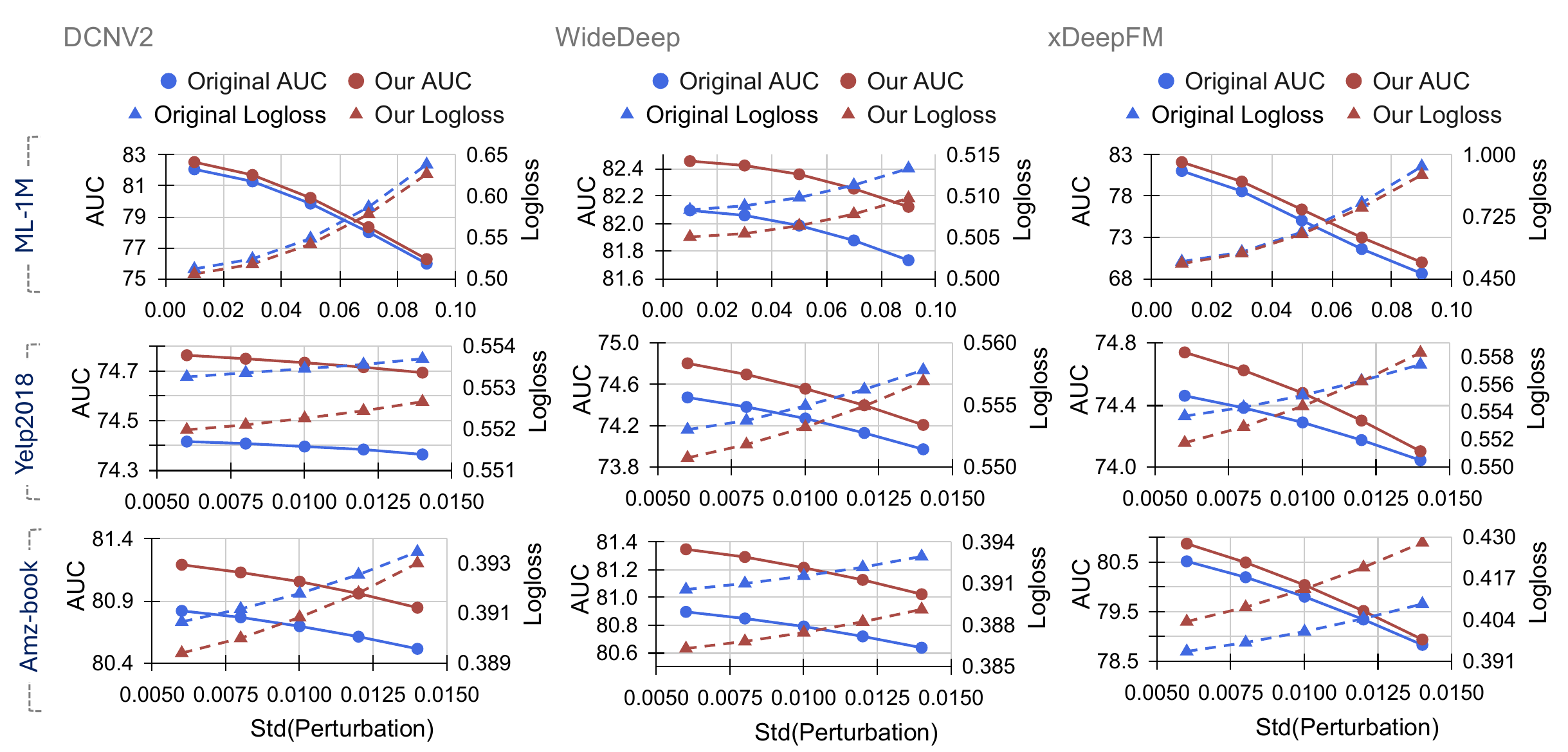}
  \caption{The changes in AUC and logloss when perturbations are applied to the input embeddings of the \textcolor[HTML]{4169E1}{\textbf{original}} CTR methods and those after applying \textcolor[HTML]{B22222}{\textbf{our}} method. Left y-axis represents AUC ($\uparrow$) and right y-axis corresponds to logloss ($\downarrow$).}
  \label{fig:perturb}
\end{figure*}

\begin{figure*}[ht]
  \centering
  \includegraphics[width=\textwidth]{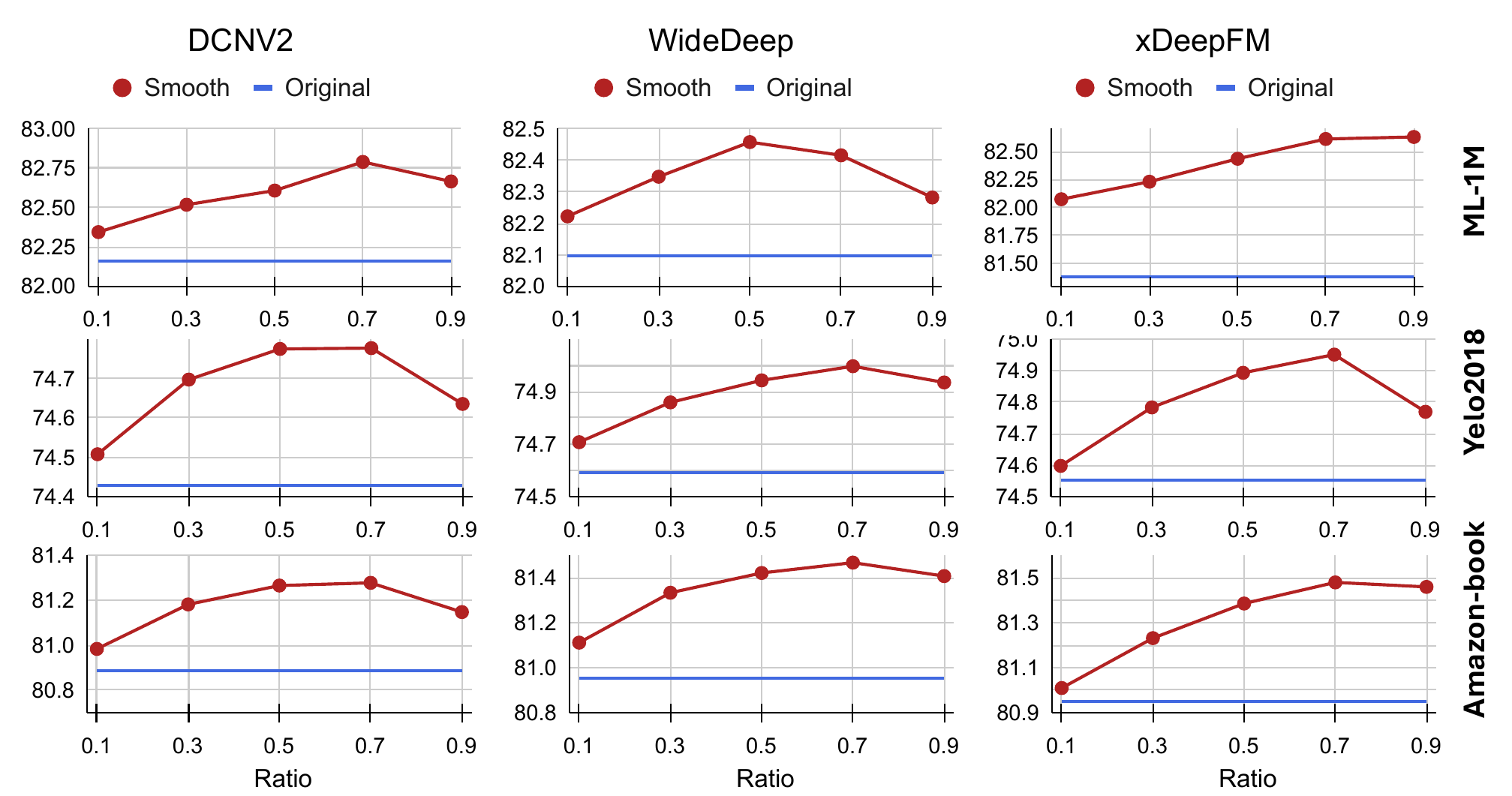}
  \caption{The AUC score of DCNV2~\cite{wang2021dcn}, WideDeep~\cite{cheng2016wide}, and xDeepFM~\cite{lian2018xdeepfm} on the ML-1M, Yelp2018, and Amazon-book dataset with different ratio $\lambda_r$ in training.}
  \label{fig:hyper}
  \vspace{-0.2in}
\end{figure*}

\subsection*{Enlarging preference supervision bandwidth as an \textit{implicit Lipschitz regularizer}}
Embeddings in recommender systems are designed to represent users and items as low-dimensional vectors that encapsulate their unique characteristics. 
Given the dynamic nature of recommendation, where user preferences and item characteristics change frequently, these embeddings should be promptly updated to capture real-time changes in user-item interaction patterns. 
Especially, frequent embedding updates help the model better adapt to rapidly evolving data, ensuring that personalized recommendations remain relevant.
However, frequent embedding updates present a challenge in optimizing both the embeddings and the subsequent learnable operations, such as MLPs and cross-layers. 
While these operations are designed to capture high-level interactions between users and items, they can be prone to overfitting. 
The fluctuations in embedding updates, driven by ever-changing user-item interactions, can cause the learnable operations to become overly sensitive to short-term data patterns, reducing their ability to generalize across the global feature space.

To address this issue, it is crucial to encourage more active and distributed adjustments in representation learning to keep up with the dynamics in recommendation.
To verify that our strategy effectively foster active representation learning, we analyze the dynamics of embedding updates during training and their properties post-training. 
Specifically, we examine the gradient norms of embeddings to assess the extent of updates made during training, providing insights into how responsive the embeddings are to input changes.
The larger the gradient norms are, the more active the embeddings are updated.
Furthermore, we analyze the weight norms of the embeddings to demonstrate the compactness and generality, which are properties critical to mitigate overfitting and ensuring model stability.
Embeddings with smaller weight norms are considered more compact and generalized.
In this experiment, we select DCNV2~\cite{wang2021dcn}, AutoInt~\cite{song2019autoint}, and FiGNN~\cite{li2019fi} as the representative CTR models and analyze their respective L2 gradient norms during training and weight norms post-training in ML-1M (relatively dense) and Yelp2018 (relatively sparse).
The results are shown in Figure~\ref{fig:grad_norm}.

From the figure, we observe: (i) compared to the respective CTR models, applying our strategy consistently results in higher gradient norms, indicating more active embedding updates during training;
(ii) The weight norms of embeddings are uniformly reduced with our approach, reflecting enhanced embedding compactness. 
Embeddings with lower weight norms exhibit less sensitivity to input noise, thereby improving model robustness and generalizability;
(iii) in contrast to the sparse dataset Yelp2018, in the dense dataset ML-1M, our method achieves a greater reduction in the embeddings' L2 gradient norms while exhibiting a smaller reduction in weight norms.
This is because dense datasets like ML-1M provide rich interaction signals, enabling our approach to leverage abundant information for more active embedding updates, resulting in numerically larger gradient norms.
Conversely, the noisy interaction signals prevalent in sparse datasets like Yelp2018 lead our method to prioritize stabilizing embeddings through reduced weight norms.

\subsection*{Implicit Lipschitz regularizer improves model robustness towards noisy perturbations}

Real-world datasets often contains varying levels of input noises.
For example, a user may adjust their rating of an item based on mainstream media reviews, or intentionally modifies their profile information to prevent privacy leakage.
A robust CTR model should be able to handle such noise without compromising its ability to capture general user preferences.
To verify that our approach improves model robustness against such noises, we introduce random Gaussian perturbations to the embeddings, with the standard deviation of the perturbations controlling the noise magnitude.
We conduct experiments using DCNV2~\cite{wang2021dcn}, WideDeep~\cite{cheng2016wide}, and xDeepFM~\cite{lian2018xdeepfm} as the representative CTR models.
For the dense dataset ML-1M, we vary standard deviation across \{0.01, 0.03, 0.05, 0.07, 0.09\};
for the sparse datasets Yelp2018 and Amazon-book, the standard deviations are chosen from \{0.006, 0.008, 0.01, 0.012, 0.014\} due to the higher sensitivity of sparse datasets to noise.
The smaller noise magnitudes for sparse datasets are necessary, as the sparsity of interactions amplifies the impact of noise to model’s generality.
We show the results in Figure~\ref{fig:perturb}.

The figure illustrates that our method exhibits superior robustness by consistently achieving higher AUCs and lower logloss values across all the selected datasets and models.
In dense datasets like ML-1M, the inherent data density reduces its sensitivity to perturbations, enabling our method to consistently outperform the original methods even when the performance has decreased by approximately 9\%.
Conversely, in sparse datasets, our method gradually loses its advantage, as sparse interactions amplify sensitivity to noisy embeddings. 
This is because more generalized embeddings may struggle to recover the original information under larger perturbations.
Nevertheless, our approach's ability to outperform the original methods under realistic noise levels in sparse datasets underscores its overall superior robustness and reliability.

\subsection*{Implicit Lipschitz regularizer achieves coefficient-invariant improvement} 
Our strategy only introduces the coefficient of the categorical cross-entropy loss $\lambda_r \in [0, 1]$ as the hyperparameter.
To verify the model's performance sensitivity to $\lambda_r$, we select DCNV2~\cite{wang2021dcn}, WideDeep~\cite{cheng2016wide}, and xDeepFM~\cite{lian2018xdeepfm} as the representative models, apply our strategy to the models with varied value for $\lambda_r$ in $\{0.1, 0.3, 0.5, 0.7, 0.9\}$, and compare the AUC performance with those of the original ones.
For each setting, we repeat the experiment under five seeds and report the average performance over the five seeds.
The comparison is shown in Figure~\ref{fig:hyper}.
From the figure, we see that our strategy consistently improves the AUC score of the original model:
with all the $\lambda_r$ values selected, our strategy yields at least 0.07\% and at most 0.65\% performance improvement.
This phenomenon suggests that our way of increasing the supervision bandwidth not only aligns well with the original CTR task, but also regularizes the learning to achieve better model generalization ability.
Moreover, while no consistent pattern emerges across models and datasets for selecting the optimal value of $\lambda_r$, the results in the figure suggest that the maximum performance gain is typically achieved when $\lambda_r$ is approximately 0.7.
Hence, in practice, initializing $\lambda_r$ at 0.7 in our method serves as a reasonable starting point, likely to deliver near-optimal performance.

\subsection*{Implicit Lipschitz regularizer enhances model generalizability with minimally increased latency}
Since our implicit regularizer configures the CTR model to produce fine-grained preference predictions in order to take advantage of the increased supervision bandwidth, we additionally examine whether/how much additional time cost is associated with the performance improvement.
We verify the extent of the time increment of our strategy by logging the time for \textit{each training epoch}, \textit{the complete training process}, and \textit{the evaluation process}.
For each benchmark dataset, we average the corresponding time for the selected CTR models in seconds and show the results in Table~\ref{tab:time}.
From the table, we see that the relative time increment is not prominent.
While it takes slightly longer to train a modified CTR model relative to the original one, the increased relative time for evaluation is rather minimal (i.e., less than $\sim 0.02\%$).
This advantage suggests our strategy's superior practicability to existing industrial recommendation pipelines, as no significant latency is introduced for prominent performance improvement.

\section*{Related Work}
\paragraph{Click-through Rate Prediction}
The problem of click-through rate (CTR) prediction is defined as predicting the interaction likelihood between a user and an item given the user ID, item ID, and optional context features.
The incorporated contextual information includes user demographic features, item descriptions, interaction timestamps, and any other information related to the user-item interaction.
These features additionally consider the variability of user preferences for items across different contexts in which they interact with the system.
Recent efforts in improving CTR prediction performance can be broadly categorized into methods focusing on feature interaction modeling, user behavior modeling, with some works exploring auxiliary methods to further enhance prediction performance.

Models focus on feature interaction aim to uncover the relationships between various input features, such as user attributes, item characteristics, and contextual factors. 
Early works like Factorization Machines (FM)~\cite{rendle2010factorization} laid the foundation for capturing low-order feature interactions efficiently by modeling pairwise relationships through factorization techniques. 
Building on this, recent methods incorporate deep learning to capture both low-order and high-order interactions. 
For example, hybrid models such as Neural Factorization Machines (NFM)~\cite{he2017neural} and DeepFM~\cite{guo2017deepfm} combine factorization techniques with deep neural networks to learn implicit and explicit feature interactions simultaneously. 
Others, like xDeepFM~\cite{lian2018xdeepfm} and Deep \& Cross Network (DCN)~\cite{wang2017deep}, introduce specialized architectures to explicitly capture feature crosses while leveraging the power of deep networks. 
Adaptive Factorization Network (AFN)~\cite{cheng2020adaptive} further extends this line of work by employing logarithmic neural transformations to adaptively learn high-order feature interactions in sparse feature spaces.
Innovations such as AutoInt~\cite{song2019autoint} and DCNV2~\cite{wang2021dcn} further refine this process by utilizing self-attention mechanisms and its variants~\cite{vaswani2017attention,mei2024esp} and improved scalability for large-scale applications.

Another significant direction focuses on understanding user interests, particularly in scenarios involving sequential or dynamic behaviors. 
These methods aim to adaptively model user preferences based on historical interactions and evolving contexts. 
Approaches like the Deep Interest Network (DIN)~\cite{zhou2018deep} and its extensions, such as the Deep Interest Evolution Network (DIEN)~\cite{zhou2019deep}, use attention mechanisms to capture the relevance of historical user actions to the current context. 
Sequential models like Deep Pattern Network (DPN)~\cite{zhang2024deep} and pretraining-based methods like SRP4CTR~\cite{han2024enhancing} emphasize learning long-term user preferences and behavior patterns to enhance personalization.

Recent research has also explored auxiliary techniques, such as contrastive learning and pretraining, to improve the generalization and robustness of CTR models. 
For example, CL4CTR~\cite{wang2023cl4ctr} introduces a contrastive learning framework to refine feature representations, while methods like SRP4CTR~\cite{han2024enhancing} leverage sequential recommendation pretraining to boost CTR prediction performance. 
Structural innovations, such as those in FinalNet~\cite{zhu2023final} and FinalMLP~\cite{mao2023finalmlp}, focus on improving training efficiency and feature interaction learning through optimized neural architectures.

Unlike the previous work, our work improves the performance of CTR prediction by shifting the attention to the bandwidth of supervision signals.
Instead of modifying the architecture or framework of the CTR models, upon the existing CTR methods focusing on feature interaction modeling, we remain their perspective model architectures and only modify the prediction heads to match the enlarged supervision through fine-grained preference signals in the data.
Such minimal structure modifications keep the model functionality intact while encouraging the model to yield smoother preference scores, which result in stronger generalizability and better performance.
\paragraph{Lipschitz Constant in Deep Models.}
The Lipschitz constant has emerged as a critical concept for understanding and controlling the stability and robustness of deep models.
Existing studies have extensively explored its applications in models leveraging convolutional and attention mechanisms~\citep{zou2019lipschitz,kim2021lipschitz,araujo2021lipschitz}. 
For example, \citet{dasoulas2021lipschitz,jia2024lipschitz, yuan2024graphmsl} propose Lipschitz normalization in graph attention networks.
\citet{das2023beyond} explores relaxing the uniform Lipschitz condition to improve optimization efficiency and privacy-utility trade-offs in the study of differential privacy.
More recently, \citet{gama2022distributed} estimated filter Lipschitz bounds using the infinity norm of matrices, contributing to stabilize graph-based neural models.
~\citet{ye2023mitigating} introduced a Lipschitz Regularized Transformer (LRFormer) to address overconfidence in transformer models within computer vision tasks. 
Despite these developments, traditional approaches to estimating Lipschitz constants often involve solving large matrix verification problems, with computational costs that increase substantially as network depth grows~\cite{xu2024eclipse}.

Our work diverges from these traditional approaches by avoiding the direct computation of Lipschitz constants. 
Instead, we utilize fine-grained preference feedback as an implicit Lipschitz regularizer in CTR models for improved robustness and generalizability.
The implicity of the Lipschitz regularizer flexibly adapts the model learning such that the smoothness of its output converges to a degree that fits specifically to the properties of trained data.

\begin{table}[ht]
\caption{The time in seconds for each training epoch, the total training process, and evaluation on NVIDIA RTX 3090 GPU. The \textcolor[HTML]{4169E1}{\textbf{Original}} row represents the original CTR model, the \textcolor[HTML]{B22222}{\textbf{Smooth}} row denotes the results after applying our approach, and the \textbf{$\Delta\%$} row shows the relative change of the time in terms of the original models'. The results are averaged over all selected CTR models.}
\centering
\begin{adjustbox}{width=\linewidth,center}
\begin{tabular}{llccc} 
\toprule
\multicolumn{2}{c}{\textbf{Time (s)}}            & \textbf{ML-1M} & \textbf{Yelp2018} & \textbf{Amazon-book}  \\ 
\midrule
\multirow{3}{*}{\textbf{Training Epoch}} & \textcolor[HTML]{4169E1}{\textbf{Original}} & 1.78  & 5.42     & 12.44        \\
                             & \textcolor[HTML]{B22222}{\textbf{Smooth}}   & 2.64  & 6.00     & 12.47        \\
                             & \textbf{$\Delta$(s)}      & 0.86 & 0.58 & 0.03         \\ 
\midrule
\multirow{3}{*}{\textbf{Full Training}}       & \textcolor[HTML]{4169E1}{\textbf{Original}} & 73.68 & 43.37    & 108.62       \\
                             & \textcolor[HTML]{B22222}{\textbf{Smooth}}   & 77.47 & 51.02    & 147.79       \\
                             & \textbf{$\Delta$(s)}      & 3.79 & 7.65 & 39.17        \\ 
\midrule
\multirow{3}{*}{\textbf{Evaluation}}  & \textcolor[HTML]{4169E1}{\textbf{Original}} & 0.12  & 0.29     & 0.48         \\
                             & \textcolor[HTML]{B22222}{\textbf{Smooth}}   & 0.13  & 0.30     & 0.50         \\
                             & \textbf{$\Delta$(s)}      & 0.01 & 0.01 & 0.02       \\
\bottomrule
\end{tabular}
\end{adjustbox}
\label{tab:time}
\end{table}

\section*{Conclusion}
In this paper, we focus on the model overfitting problem in the context of CTR prediction.
We first point out that most CTR models, especially their embeddings, cannot keep pace with the fast-changing recommendation environment where user preferences and item characteristics change frequently.
As a result, they inevitably overfit the short-term fluctuations in the data rather than the high-level interactions.
Then theoretically, we discover that increasing the supervision bandwidth reduces the model's Lipschitz constant, which inversely measures a model's stability with the input.
Empirically, following the above theorem, we propose to enlarge the supervision bandwidth by recovering the refined rating preferences during the model training process in addition to the existing binary supervisions.
Through extensive experiments, we verify that our strategy:
(i) generally improves both the predictive and rating performance of CTR models;
(ii) decouples the learning of embeddings and the learnable operations (e.g., MLPs, cross layers, etc.) to alleviate overfitting;
(iii) achieves consistent performance improvement with any reasonable coefficient;
(iv) exhibits superior practicability with minimal extra evaluation latency.

\section*{Limitations}
While our study provides both theoretical and empirical evidence that increasing supervision bandwidth enhances model stability and generalizability, we recognize several limitations:
(i) each CTR model possesses its own distinct architecture. Determining the most effective way to modify these structures to support the expanded supervision scheme is an open problem, as there may be multiple approaches to adjust each CTR model. Developing comprehensive guidelines tailored to these architectural adaptations is crucial for fully harnessing the benefits of augmented supervision and achieving optimal model performance.
Nevertheless, our current implementation of modifying only prediction layers only is simple yet effective.
(ii) expanding the supervision bandwidth necessitates access to more granular preference feedback. However, identifying fine-grained ground truth becomes challenging in contexts where such detailed feedback is not readily available. Interaction metrics such as clicks, dwell time, and other forms of engagement do not inherently convey the nuances of user preferences. Therefore, further research is needed to explore methods for applying our strategy to CTR models trained primarily on implicit feedback, identifying alternative ways to infer and incorporate fine-grained preferences.
We extend the discussion in Appendix~\ref{appd:availability}.

\section*{Acknowledgments}
This research was supported in part by the National Science Foundation under Grant No. 2452367.

\bibliography{aaai25.bib}

\section*{Paper Checklist}
\begin{enumerate}
\item For most authors...
    \begin{enumerate}
        \item  Would answering this research question advance science without violating social contracts, such as violating privacy norms, perpetuating unfair profiling, exacerbating the socio-economic divide, or implying disrespect to societies or cultures?
        \answerYes{Yes, and our research aligns with the core principles of ethical research and social responsibility.}
      \item Do your main claims in the abstract and introduction accurately reflect the paper's contributions and scope?
        \answerYes{Yes}
       \item Do you clarify how the proposed methodological approach is appropriate for the claims made? 
        \answerYes{Yes}
       \item Do you clarify what are possible artifacts in the data used, given population-specific distributions?
       \answerNo{No, because the data used in this work reflects personal preferences.}
      \item Did you describe the limitations of your work?
        \answerYes{Yes, in the Limitations section.}
      \item Did you discuss any potential negative societal impacts of your work?
        \answerNo{No, because this work mainly discuss how to improve model generalizability, which leads to more accurate user preference prediction.}
          \item Did you discuss any potential misuse of your work?
        \answerNo{No, because the work is intended to improve model generalizability and stability.}
        \item Did you describe steps taken to prevent or mitigate potential negative outcomes of the research, such as data and model documentation, data anonymization, responsible release, access control, and the reproducibility of findings?
        \answerNo{No, because the data does not contain personalized information other than the interaction histories. The users and items are anonymized.}
      \item Have you read the ethics review guidelines and ensured that your paper conforms to them?
        \answerYes{Yes.}
    \end{enumerate}

\item Additionally, if your study involves hypotheses testing...
    \begin{enumerate}
      \item Did you clearly state the assumptions underlying all theoretical results?
        \answerNA{NA}
      \item Have you provided justifications for all theoretical results?
        \answerNA{NA}
      \item Did you discuss competing hypotheses or theories that might challenge or complement your theoretical results?
        \answerNA{NA}
      \item Have you considered alternative mechanisms or explanations that might account for the same outcomes observed in your study?
        \answerNA{NA}
      \item Did you address potential biases or limitations in your theoretical framework?
        \answerNA{NA}
      \item Have you related your theoretical results to the existing literature in social science?
        \answerNA{NA}
      \item Did you discuss the implications of your theoretical results for policy, practice, or further research in the social science domain?
       \answerNA{NA}
    \end{enumerate}
    
\item Additionally, if you are including theoretical proofs...
    \begin{enumerate}
      \item Did you state the full set of assumptions of all theoretical results?
        \answerNA{NA}
    	\item Did you include complete proofs of all theoretical results?
        \answerNA{NA}
    \end{enumerate}

\item Additionally, if you ran machine learning experiments...
    \begin{enumerate}
      \item Did you include the code, data, and instructions needed to reproduce the main experimental results (either in the supplemental material or as a URL)?
        \answerYes{Yes, the data is public and the codes and are provided in the anonymous URL.}
      \item Did you specify all the training details (e.g., data splits, hyperparameters, how they were chosen)?
        \answerYes{Yes, in section Experimental Settings.}
         \item Did you report error bars (e.g., with respect to the random seed after running experiments multiple times)?
        \answerNo{No, because the variances are minimal and we omit it following previous works.}
    	\item Did you include the total amount of compute and the type of resources used (e.g., type of GPUs, internal cluster, or cloud provider)?
        \answerYes{Yes, in section Experimental Settings.}
         \item Do you justify how the proposed evaluation is sufficient and appropriate to the claims made? 
        \answerYes{Yes, in section Comparison with CL-based Methods.}
         \item Do you discuss what is ``the cost`` of misclassification and fault (in)tolerance?
        \answerNo{No, because under the case of recommendation, there is not an absolutely right or wrong decision.}
      
    \end{enumerate}

\item Additionally, if you are using existing assets (e.g., code, data, models) or curating/releasing new assets, \textbf{without compromising anonymity}...
    \begin{enumerate}
      \item If your work uses existing assets, did you cite the creators?
        \answerYes{Yes, in section Experimental Settings.}
      \item Did you mention the license of the assets?
        \answerYes{Yes, the licenses are mentioned in the cited papers.}
      \item Did you include any new assets in the supplemental material or as a URL?
        \answerYes{Yes, we provide it in an anonymous URL.}
      \item Did you discuss whether and how consent was obtained from people whose data you're using/curating?
        \answerNo{No, because the data is publicly available for research purposes.}
      \item Did you discuss whether the data you are using/curating contains personally identifiable information or offensive content?
        \answerNo{No, because the data is anonymized.}
    \item If you are curating or releasing new datasets, did you discuss how you intend to make your datasets FAIR?
    \answerNA{NA}
    \item If you are curating or releasing new datasets, did you create a Datasheet for the Dataset? 
    \answerNA{NA}
    \end{enumerate}

\item Additionally, if you used crowdsourcing or conducted research with human subjects, \textbf{without compromising anonymity}...
    \begin{enumerate}
      \item Did you include the full text of instructions given to participants and screenshots?
        \answerNA{NA}
      \item Did you describe any potential participant risks, with mentions of Institutional Review Board (IRB) approvals?
        \answerNA{NA}
      \item Did you include the estimated hourly wage paid to participants and the total amount spent on participant compensation?
        \answerNA{NA}
       \item Did you discuss how data is stored, shared, and deidentified?
       \answerNA{NA}
    \end{enumerate}
\end{enumerate}

\appendix

\section{Discussions}

\subsection{Explicit Preference Feedback Availability}\label{appd:availability}
While our method requires explicit preference feedback to leverage its full potential, the availability of such feedback is not a significant barrier. 
Many widely used datasets already provide fine-grained preference feedback in the form of ratings or similar metrics. 
Examples include but not limit to Movielens\footnote{\url{https://grouplens.org/datasets/movielens/}}, Book-Crossing\footnote{https://grouplens.org/datasets/book-crossing/}, Anime\footnote{https://www.kaggle.com/datasets/CooperUnion/anime-recommendations-database}. 
All of the datasets are extensively used in assessing research studies in recommender systems~\cite{wang2019knowledge,lin2024clickprompt}.
Even in cases where fine-grained feedback is unavailable, reasonable proxies are often available and can be employed. 
For instance, the Frappe\footnote{http://baltrunas.info/research-menu/frappe} dataset includes information on the number of times an app has been used by a user, where a larger number of usage times indicates a stronger preference. 
Similarly, the Last.FM\footnote{https://grouplens.org/datasets/hetrec-2011/} dataset provides listening counts, which are indicative of user preferences. 
These proxy statistics, while not explicit ratings, offer sufficient granularity to be utilized as a substitute for direct feedback.

Lastly, it is crucial to reiterate the broader aim of this work: addressing the limitations of current practices in processing fine-grained feedback in state-of-the-art CTR prediction models. 
Many existing methods~\cite{wang2019knowledge,lin2024clickprompt} rely on thresholding to convert fine-grained feedback into binary labels, inherently losing valuable preference information. 
This paper advocates for an alternative approach to better utilize fine-grained feedback, emphasizing the need for a paradigm shift in how such data is perceived and processed. 
Encouraging the collection and use of more fine-grained feedback could lead to significant advancements in the field.

\subsection{Beyond CTR Prediction}\label{appd:beyondCTR}
Our method addresses the issue of overfitting in CTR prediction by leveraging fine-grained preference feedback, which is often underutilized by recent works due to limitations introduced by thresholding. 
This approach is specifically designed to accommodate the dynamic nature of recommendation and is tailored for binary classification tasks where such information loss prominently contributes to overfitting.
For recommendation tasks that also involve binary classification and suffer from the same loss of information due to thresholding, our method is expected to be directly applicable. 
However, for tasks that do not meet these criteria, modifications to the method would be necessary to align with the task's unique requirements.
At its core, our method's strength lies in its focus on stabilizing models through increased supervision bandwidth. 
This principle is not limited to CTR prediction and can be adapted to other suitable tasks, provided that the method is properly modified to address the specific challenges of those tasks. 
This adaptability underscores the broader applicability of the approach to enhancing model stability across various recommendation scenarios.

\subsection{Applicability to Other Structures}\label{appd:other_structures}
The theoretical foundation of increasing supervision bandwidth as a regularization method suggests it is broadly applicable to various neural network architectures, including those with specialized structures or non-ReLU activation functions, such as Mamba~\cite{gu2024mamba}, KAN~\cite{qiu2024relu}, TTT~\cite{sun2024learning}, and boarder GNN-based methods~\cite{}. The core idea hinges on stabilizing the learning process by controlling the Lipschitz constant, which is not inherently tied to a specific activation function or network structure.
However, the effectiveness of this approach in such networks depends on how supervision bandwidth interacts with the specific characteristics of these architectures, such as their activation dynamics and structural constraints. 
For instance, non-ReLU activations might exhibit different sensitivities to overfitting or instability, requiring adjustments to the supervision strategy.
In principle, the method is adaptable, but practical application to these architectures would require empirical validation and potentially task-specific modifications to optimize performance while maintaining theoretical guarantees.
\begin{figure*}[ht]
  \centering
  \includegraphics[width=\textwidth]{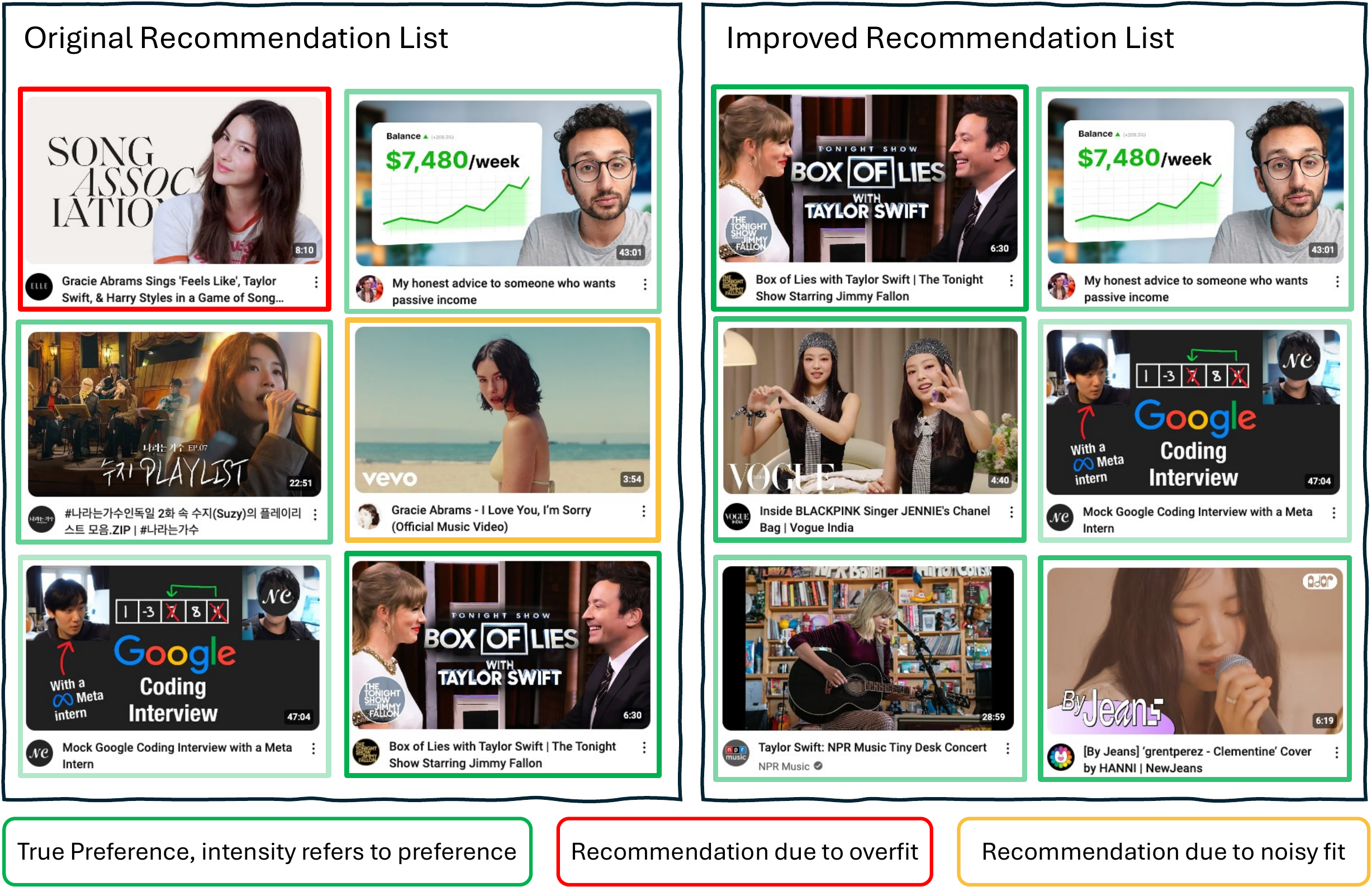}
  \caption{A concrete example for end user experience improvement in recommendation. Compared with the recommendation list generated by the original model (left side), with improved generality, our models generates recommended items (right side) that is both accurate (i.e., contents are preferred) and personalized (i.e., more interested items are ranked in the front).}
  \label{fig:end_user}
\end{figure*}

\subsection{From the End-user Experience}\label{appd:end_user}
Our study demonstrates that leveraging fine-grained supervision signals enhances not only ranking metrics, such as NDCG and Recall, but also predictive metrics in CTR models. 
These improvements translate directly into better user experiences across various stages of interaction with recommendation systems:
(i) higher ranking metrics ensure that the \textit{most relevant items consistently appear at the top of recommendation lists}, fostering trust and reducing cognitive effort during decision-making. 
For example, users browsing for recommended movies are more likely to find their preferred movies prominently displayed, creating an efficient and satisfying movie recommendation experience. 
Improved Recall further broadens the range of relevant items presented, enabling users to discover items they might not have explicitly demonstrate preferences, such as items liked by similar users. 
This enhances engagement and supports serendipitous discoveries, enriching the overall user journey;
(ii) In addition to ranking, the improved predictive metrics ensure that the \textit{recommendations more accurately reflect user preferences}, leading to greater personalization. 
For instance, with a higher CTR prediction rate, the system can prioritize items that users are more likely to engage with. 
This refinement improves the ranking compared to the original model by making the top-listed items not only relevant but also highly actionable, reducing irrelevant or marginally useful recommendations.
By enhancing both predictive and ranking metrics, our method supports the overarching goal of delivering recommendation systems that perform well quantitatively while providing meaningful, satisfying, and personalized user experiences.
We show a concrete example for the enhanced user experience in Figure~\ref{fig:end_user}.

\subsection{Stabilizing Relational Learning and their Applications: Beyond the Lipschitz Methods}
Stability in relational learning models, especially in GNN and their application in recommendation systems, has been studied from multiple perspectives. A prominent direction involves explicitly regularizing the Lipschitz constant to enhance robustness and generalizability. For instance, \citet{yuan2024mitigating} identifies an emergent robustness degradation phenomenon in graph representation learning when models are scaled up, addressing it through a generalized knowledge distillation framework. Similarly, \citet{wen2024gcvr} improves graph contrastive learning by reconstructing representations from cross-view information, implicitly stabilizing representations against data perturbations. \citet{zhang2023sparsity} further demonstrates that sparsity-driven contrastive models not only reduce data requirements but also enhance representation stability, addressing class imbalance problems. \citet{zhang2022look} generalizes the relational learning to the image captioning tasks, which uses self-supervised contrastive learning to stabilize the image captioning training. \citet{qian2022co} uses self-supervised contrastive learning to stably learn to process relational data such as AMiner. \citet{tian2023hgm} explores the mask autoencoder's effect to ease the relational graph model's training. 

Several other approaches have directly aimed at enhancing all-around robustness in graph learning through model architecture, training methods, and optimization techniques. \citet{zhang2023chasing} presents an integrated framework for robust graph representation learning by comprehensively modeling and mitigating various robustness risks. Meanwhile, \citet{zhang2023gap} tackles the critical challenge of distribution gaps in graph few-shot learning, showing improved model stability and performance by explicitly bridging these gaps. \citet{liu2023fair} extends fairness concerns into a unified framework, proposing diverse mixture-of-experts architectures to learn fair and stable graph representations, which have broader societal implications.

Alongside these explicit Lipschitz and architectural methods, recent research also leveraged knowledge distillation to stably improve the generalization of GNNs. \citet{guo2023boosting} proposes an adaptive knowledge distillation approach, refining the learning objectives of student models through teacher-student interactions, thus improving GNN generalization and robustness. Similarly, \citet{yue2022label} employs label-invariant data augmentation strategies, enhancing semi-supervised graph classification performance and robustness.

\section{Proof of Theorem~\ref{thm-1}}\label{appd:proof}
\begin{proof}
Let $\mathbf{f}(\mathbf{x}) = [f_1(\mathbf{x}), \dots, f_N(\mathbf{x})]$ be the logits output by the model where $N$ is the dimension of the output logits. 
The softmax normalization function with the built-in temperature scaling is defined as:
\begin{equation}
p_i(\mathbf{x}) = \frac{\exp\left(f_i(\mathbf{x})/\tau\right)}{\sum_{j=1}^{N} \exp\left(f_j(\mathbf{x})/\tau\right)} \quad \text{for } i = 1, 2, \dots, N,
\end{equation}
where $\tau > 0$ is the temperature parameter that adjusts the uniform distribution.
We aim to derive a relation on the Lipschitz constant $L_p(N)$ of the temperature-scaled probabilities with respect to the input $\mathbf{x}$. The Lipschitz constant is defined as the smallest $L_p(N)$ such that for all $\mathbf{x}, \mathbf{y} \in \mathcal{X}$,
\begin{equation}
\|\mathbf{p}(\mathbf{x}) - \mathbf{p}(\mathbf{y})\| \leq L_p(N) \|\mathbf{x} - \mathbf{y}\|.
\end{equation}
The sensitivity of the model can be analyzed by considering the Jacobian matrix of the model output probabilities with the output logits. The Jacobian $\mathbf{J_{\sigma}}(\mathbf{f})$ is an $N \times N$ matrix:
\begin{equation}
\frac{\partial p_i(\mathbf{x})}{\partial f_j(\mathbf{x})} = p_i(\mathbf{x}) (\delta_{ij} - p_j(\mathbf{x})),
\end{equation}
where $\delta_{ij}$ is the Kronecker delta. The Frobenius norm of this Jacobian matrix provides a measure of the sensitivity of the model output to changes in the logits:
\begin{equation}
\|\mathbf{J_{\sigma}}(\mathbf{f})\|_F = \sqrt{\sum_{i=1}^{N} \sum_{j=1}^{N} \left( p_i(\mathbf{x}) (\delta_{ij} - p_j(\mathbf{x})) \right)^2}.
\end{equation}
When the logits are evenly distributed as $N$ is increasing, $p_i$ approaches $\frac{1}{N}$ for all $i$, aligning closely with insights from label smoothing~\cite{muller2019does}. 
Then, the elements of the Jacobian matrix can be estimated as:
\begin{equation}
\frac{\partial p_i}{\partial f_j} \approx \frac{1}{N} (\delta_{ij} - \frac{1}{N}),
\end{equation}
and the Frobenius norm becomes:
\begin{align}
\|\mathbf{J_{\sigma}}(\mathbf{f})\|_F &\approx \sqrt{N \cdot \left(\frac{1}{N} \left(1-\frac{1}{N}\right)\right)^{2}} \notag \\
&\quad + N(N-1) \cdot \left(\frac{1}{N^2}\right)^2.
\end{align}
The dominant term for large $N$ is:
\begin{equation}
\|\mathbf{J_{\sigma}}(\mathbf{f})\|_F \approx \frac{1}{\sqrt{N}}.
\end{equation}
Given that the logits $\mathbf{f}(\mathbf{x})$ are Lipschitz continuous with Lipschitz constant $L_f$, the sensitivity of the model output with respect to the input $\mathbf{x}$ can be constrained by:
\begin{align}
\|\mathbf{p}(\mathbf{x}) - \mathbf{p}(\mathbf{y})\| &\leq \|\mathbf{J_{\sigma}}(\mathbf{f})\|_F \cdot \|\mathbf{f}(\mathbf{x}) - \mathbf{f}(\mathbf{y})\| \notag \\
&\leq \|\mathbf{J_{\sigma}}(\mathbf{f})\|_F \cdot L_f \|\mathbf{x} - \mathbf{y}\|.
\end{align}
The Frobenius norm of the model Jacobian matrix is constrained by:
\begin{equation}
\|\mathbf{J_{\sigma}}(\mathbf{f})\|_F \leq \frac{1}{\sqrt{N}},
\end{equation}
as each component $p_i(\mathbf{x}) (\delta_{ij} - p_j(\mathbf{x}))$ is small when the logits are comparable, and the summation across $N$ terms distributes the influence~\cite{miyato2018spectral}.
Thus, the Lipschitz constant $L_p(N)$ of the model function satisfies:
\begin{equation}
L_p(N) \leq \frac{L_f}{\sqrt{N}}.
\end{equation}
\end{proof}

\section*{Data Statistics}\label{appd:data_stats}
The statistics of the benchmark datasets are shown in Table~\ref{tab:data}.
\begin{table}[h]
\caption{The statistics of the benchmark datasets.}
\label{tab:data}
\begin{adjustbox}{width=\linewidth,center}
\centering
\begin{tabular}{lrrrc} 
\toprule
\textbf{Dataset}      & \textbf{\#User} & \textbf{\#Item} & \textbf{\#Interaction} & \textbf{Sparsity}  \\ 
\midrule
ML-1M        & 6,041    & 3,261    & 998,539         & 0.9493    \\
Yelp2018     & 77,278   & 45,639   & 2,103,896        & 0.9994    \\
Amazon-Book & 68,498   & 65,549   & 2,954,716        & 0.9993    \\
\bottomrule
\end{tabular}
\end{adjustbox}
\end{table}

\end{document}